# The ORCA Benchmark: Evaluating Real-World Calculation Accuracy in Large Language Models

Claudia Herambourg[1], Dawid Siuda[2], Anna Szczepanek[2], Julia Kopczyńska[3], Joao R. L. Santos[4], Wojciech Sas[2], Joanna Śmietańska-Nowak[2]

[1]Department of Linguistics and Phonetics (ILGPA), Sorbonne Nouvelle University, Paris, France

[2]Omni Calculator Research Team

[3]Institute of Biochemistry and Biophysics, Polish Academy of Sciences, Warsaw, Poland

[4]UFCG - Universidade Federal de Campina Grande, Unidade Acadêmica de Física, 58429-900 Campina Grande, PB, Brazil, Institute for Theoretical Physics, Leibniz University Hannover, Appelstraße 2, 30167 Hannover, Germany

Corresponding authors: dawid.siuda@omnicalculator.com, claudia.herambourg@omnicalculator.com, joanna.smietanska@omnicalculator.com

**Abstract**

We present ORCA (Omni Research on Calculation in AI) Benchmark - a novel benchmark that evaluates large language models (LLMs) on multi-domain, real-life quantitative reasoning using verified outputs from Omni's calculator engine. In 500 natural-language tasks across domains such as finance, physics, health, and statistics, the five state-of-the-art systems (ChatGPT-5, Gemini 2.5 Flash, Claude Sonnet 4.5, Grok 4, DeepSeek V3.2) achieved only 45–63% accuracy, with errors mainly related to rounding (35%) and calculation mistakes (33%). Results in specific domains indicate strengths in mathematics and engineering, but weaknesses in physics and natural sciences. Correlation analysis with r ≈ 0.40–0.65 shows that the models often fail together but differ in the types of errors they make, highlighting their partial complementarity rather than redundancy. Unlike standard math datasets, ORCA evaluates step-by-step reasoning, numerical precision, and domain generalization across real problems from finance, physics, health, and statistics.

## 1. Introduction

The rapid evolution of large language models (LLMs) has profoundly transformed the landscape of artificial intelligence. Systems such as ChatGPT-5 [1], Gemini 2.5 Flash [2], Claude 4.5 Sonnet [3], and Grok 4 [4] now display impressive capabilities in natural language comprehension, reasoning, and problem-solving across a vast range of domains. This rapid progress has sparked both excitement and critical inquiry into the depth and reliability of LLM reasoning, particularly in one of the most demanding domains of human cognition: mathematical thinking [5], [6].

Evaluating the mathematical capacity of LLMs is crucial for understanding their reasoning limits [7]. Existing benchmarks have already covered a wide spectrum of mathematical challenges. For instance, GSM8K [6] focuses on middle-school-level word problems, MATH [8] targets high-school competition problems, while ProofNet [9] and OCWCourses [10] address university-level reasoning. At the upper end, MiniF2F [11] and AlphaGeometry [12] provide Olympiad-grade problems. Collectively, these benchmarks have become essential tools for measuring mathematical proficiency. Yet they predominantly assess academic or formal mathematics, such as symbolic derivations, proofs, or abstract reasoning, rather than the type of practical, applied mathematics that people use in everyday life [13].

Despite their remarkable achievements, LLMs continue to exhibit fundamental limitations across multiple dimensions of cognition [14]. They often misinterpret linguistic nuance, struggle with commonsense and contextual understanding, and lack embodied or spatial awareness [15]. Their reasoning remains fragile in areas such as relational and logical inference, where intuition or real-world grounding is required. Even in mathematics, models frequently rely on surface pattern matching rather than rule-based calculation, producing brittle and inconsistent results [14], [16]. Moreover, they tend to propagate misinformation inherited from their training data and exhibit overfitting, performing well on familiar benchmarks but failing to generalize to unseen or complex problems [17].

Although models such as GPT-4 have demonstrated outstanding accuracy on datasets like GSM8K (over 90%) and MATH (around 80%), prior research [8], [18] shows that they remain prone to logical inconsistencies and cascading arithmetic errors in multi-step reasoning [19]. Furthermore, many of these datasets may have been partially included in model training corpora, potentially inflating reported scores. This raises the need for fresh, domain-specific evaluations grounded in problems that models have not seen before and that reflect genuine computational reasoning rather than memorized patterns.

To address precisely these gaps, we propose a new ORCA (Omni Research on Calculation in AI) Benchmark designed to assess the ability of modern language models to solve real-world calculation tasks. We aim to evaluate how effectively the newest, popular LLMs can perform real-world, everyday calculations - the kind that real users perform using our verified calculator engines. These include problems from fields like finance, physics, health, and statistics, where accuracy is critical and where solutions must follow clear, structured reasoning rather than rely on linguistic plausibility. Unlike existing mathematical benchmarks that test competition-style reasoning or abstract algebraic manipulation, our framework examines how well LLMs reproduce correct, step-by-step quantitative reasoning in daily-use contexts.

In our study, we systematically test five leading models: ChatGPT-5, Gemini 2.5 Flash, Claude 4.5 Sonnet, Grok 4-on a wide set of calculator-based tasks. Each model receives standardized prompts equivalent to the data fed to our own engines, and its responses are automatically compared to verified results. The outcome is a simple but powerful measure: the percentage of correctly solved problems (e.g., *ChatGPT-5: 50%, Gemini 4.5: 63%*). This approach reveals how closely LLMs can approximate deterministic, verifiable computation - an essential capability for any model aspiring to reliability in applied reasoning.

In summary, our work bridges the gap between language-based reasoning and real-world computational accuracy. By grounding evaluation in authentic calculator tasks, it transforms practical mathematical competence into a measurable dimension of intelligence - revealing not only how advanced today's models have become, but also how far they still are from mastering the kind of reasoning humans use every day.

## 2. Background and Related Work

### 2.1. Limitations of LLM Reasoning Benchmarks

Evaluating the mathematical and logical reasoning capacity of large language models (LLMs) remains a central challenge in understanding their true cognitive capabilities. Existing benchmarks, while foundational, remain narrow in scope and fail to capture the full spectrum of reasoning robustness. Early math word problem (MWP) datasets, such as those introduced by Kushman et al. (2014) [20] and Roy and Roth (2015) [21], were relatively small and primarily designed for traditional machine learning models. Consequently, they provide only a limited view of LLM reasoning performance. Later efforts, including Ape210K [22], expanded dataset size but not depth: its elementary-school focus, linguistic barrier, and lack of natural-language reasoning steps restrict its applicability to modern, English-based models.

Standard benchmarks such as GSM8K [6] and MATH [8] advanced the field by introducing high-quality, linguistically diverse problems, yet they primarily assess procedural competence rather than higher-order reasoning. Evaluating tasks that demand theoretical understanding, creative insight, or multi-step symbolic derivation remains an open challenge [17]. A similar limitation persists in logical reasoning benchmarks, which tend to emphasize formal inference patterns over everyday contextual understanding [23].

Another persistent issue is data contamination - the inadvertent inclusion of benchmark problems or close variants in model training corpora [24], [25]. This artificially inflates evaluation results and obscures genuine generalization ability. Golchin and Surdeanu [26], for example, identified such contamination in GPT-4 on GSM8K-style tasks under guided prompting conditions. Beyond numerical inaccuracies, evidence of hallucination-related reasoning errors in recent reasoning-oriented LLMs (RLLMs) were also observed [27]. The analysis revealed that popular models (e.g., OpenAI's o1-mini and o3-mini, DeepSeek-R1, Claude 3.7 Sonnet, Gemini 2.5 Pro Preview, and Grok 3 Mini Beta) frequently hallucinate edges not specified in the prompt's graph description.

While recent research trends push benchmarks toward increasingly advanced mathematical challenges [5], [13], [16], [28], [29], they often overlook another critical dimension: reliability in simple arithmetic and elementary quantitative reasoning. This persistent fragility indicates that progress in complex reasoning does not necessarily translate into dependable computational accuracy, underscoring the need for benchmarks that test both advanced reasoning and basic numerical robustness.

### 2.2. Model Selection and Mathematical Reasoning Capabilities

Our study evaluates the mathematical problem-solving capabilities of five representative large language models: GPT-5, DeepSeek V3.2, Gemini 2.5 Flash, Claude 4.5 Sonnet, and Grok 4. Importantly, this selection includes the free, most advanced, and widely used models available during the evaluation period in October 2025, ensuring that the benchmark captures both state-of-the-art proprietary systems and publicly accessible alternatives. This choice allows for a balanced assessment of LLM mathematical reasoning across different accessibility levels and computational paradigms.

#### 2.2.1. ChatGPT-5

GPT, developed by OpenAI, is a large language model based on the Transformer architecture [30], [31], which uses a self-attention mechanism to capture long-range dependencies and structural relationships in text. This allows it to interpret mathematical expressions and perform complex symbolic and numerical reasoning. Pre-trained on large-scale corpora through self-supervised learning, GPT acquires broad linguistic competence and general reasoning skills. It is further refined using Reinforcement Learning from Human Feedback (RLHF) [32], aligning its responses with human expectations for logic, clarity, and problem-solving style. GPT demonstrates strong performance in algebraic manipulation, equation solving, function analysis, and limit evaluation, often providing clear, step-by-step derivations [19]. It also supports applied modelling in probability, statistics, and optimization, and can generate executable code (Python, MATLAB, R, Stata) for computational analysis. Additionally, GPT effectively translates natural language problems into formal mathematical representations and offers structured explanations of advanced topics such as linear algebra or differential equations, making it a valuable tool for reasoning, teaching, and research in mathematical contexts [33].

### 2.2.2. Gemini 2.5 Flash

Gemini 2.5 Flash, developed by Google DeepMind, belongs to the multimodal Gemini 2 family of models, designed to integrate reasoning across text, code, and visual modalities. Built upon the Transformer framework, Gemini incorporates architectural and training refinements that emphasize efficiency, contextual compression, and long-context reasoning. The "Flash" variant is optimized for high-speed inference and lightweight deployment, maintaining strong reasoning performance while significantly reducing latency and computational cost [28]. Like other Gemini models, it is trained on large-scale, multimodal datasets combining natural language, code, and mathematical content, which enhances its symbolic and quantitative reasoning abilities. In mathematical domains, Gemini 2.5 Flash demonstrates strong results on multistep reasoning, code-assisted problem-solving, and symbolic manipulation, benefitting from its integrated programming capabilities and instruction-tuned alignment [34]. It also shows robust performance in structured quantitative tasks, such as algebraic simplification and numerical modelling, making it well-suited for applied reasoning evaluations.

### 2.2.3. Claude 4.5 Sonnet

Claude 4.5 Sonnet, developed by Anthropic, is an LLM oriented toward reasoning, mathematics, coding, and tool-mediated problem-solving. Built on the Transformer architecture, it incorporates extensive fine-tuning and reinforcement learning to strengthen multistep reasoning and improve contextual precision. The model was trained on a proprietary dataset that blends public, licensed, and user-contributed material available up to July 2025 [35]. Anthropic implemented comprehensive data cleaning, filtering, and alignment procedures, guided by Reinforcement Learning from Human and AI Feedback (RLHF/RLAIF), to support its "helpful, honest, and harmless" (HHH) alignment framework. Empirical evaluations indicate that Claude 4.5 delivers high factual accuracy and strong reasoning depth, though its output tends to be linguistically dense [36]. While it performs reliably in structured reasoning tasks, studies [37] suggest it can occasionally obscure its chain of thought when exposed to adversarial or high-pressure prompts, underscoring the importance of transparent reasoning monitoring. In summary, Claude 4.5 balances advanced analytical performance with a rigorous commitment to safety and interpretability, standing out as one of the most capable and carefully aligned reasoning models of its generation.

### 2.2.4. Grok 4

Grok 4, the newest product from xAI, remains a reasoning-focused LLM designed for advanced mathematics, structured problem solving, and collaborative reasoning. It incorporates long-context attention and alignment refinements to improve interpretability and consistency across multi-step reasoning tasks. Empirical evaluations position Grok 4 near the state of the art in mathematical reasoning. It performs exceptionally well on medium-to-hard high school competition problems and ranks among the strongest models on proof-based tasks, though substantial headroom remains for fully rigorous proofs. In the IMProofBench evaluation [38], Grok 4 achieved the highest accuracy (52%) on final-answer subproblems, indicating reliable quantitative reasoning but lingering challenges in open-ended theoretical domains. Professional mathematicians have noted that Grok 4 may currently be the best available model for mathematical literature search, efficiently retrieving and contextualizing formal results [39]. Interestingly, the model shows a tendency to detect some of its own reasoning errors, though not consistently: a feature that hints at early forms of metacognitive behavior. Beyond individual reasoning, Grok 4 has also shown promising results in multi-agent orchestration frameworks [40], where multiple LLMs interact and vote iteratively to reach consensus. Nonetheless, it remains xAI's most capable reasoning system to date, combining analytical precision with emerging self-correction abilities and collaborative reasoning potential.

### 2.2.5. DeepSeek V3.2

DeepSeek V3.2, developed by DeepSeek AI in China, is a large language model built on a refined Mixture-of-Experts (MoE) architecture that balances computational efficiency with high reasoning capacity. The model employs Multi-head Latent Attention (MLA) and the DeepSeekMoE framework, activating 37 billion of its 671 billion parameters per token [41]. This design enables strong performance while maintaining efficient inference and stable training. The model shows particular strength in mathematical modelling, symbolic computation, and scientific writing, with results approaching those of closed-source systems such as GPT-4 Turbo [42]. DeepSeek V3.2 also demonstrates solid chain-of-thought reasoning and integrates seamlessly with companion modules like DeepSeek-Coder and DeepSeek-VL, supporting mixed-media mathematical tasks that combine text, code, and visual data. While its structured reasoning and cross-lingual consistency mark a major leap over earlier DeepSeek models, challenges remain in handling open-ended reasoning and adaptive problem-solving.

## 3. Methodology

### 3.1. Dataset

The final ORCA benchmark dataset comprises 500 prompts spanning 13 categories represented by our calculators from the Omni Calculator website [43]: **Math, Finance, Health, Everyday Life, Construction, Physics, Biology, Chemistry, Conversion, Ecology, Sports, Statistics**, and **Other**. This broad coverage distinguishes our benchmark, as it reflects the full range of everyday quantitative problems faced by regular users rather than narrow academic or competition tasks. Each prompt corresponds directly to an existing Omni calculator, ensuring that every problem has a single deterministic ground-truth answer verified by our computation engine and experts. Prompts were designed to reflect realistic user intentions and natural phrasing, simulating how everyday users would articulate their questions (e.g., "If I deposit $50,000 at 5% APR, compounded weekly, what will my balance be after 18 months?"). To capture a range of reasoning demands, each question was classified into one of three difficulty levels - Easy, Medium, or Hard - based on the number of reasoning steps, required formulas, and potential for computational or unit-based errors. The exemplary set of prompts, reference answers, and possible error types is provided in the Appendix.

## 3.2. Expert Recruitment and Prompt Development

Prompt creation was conducted by domain experts, consisting of Omni researchers with academic and professional backgrounds in relevant disciplines. For example, prompts in physics were authored by contributors holding PhDs in physics or engineering, while specialists in their respective fields designed financial and health-related prompts. This expert-based recruitment process ensured that the prompts accurately represented authentic, discipline-specific reasoning tasks, preserving both conceptual rigor and real-world relevance. Each expert maintained a consistent approach to prompt formulation to ensure stable structure and comparable difficulty across domains.

## 3.3. Benchmark Design

Each prompt was tested against five newest, leading LLMs: ChatGPT-5, Gemini 2.5 Flash, Claude 4.5 Sonnet, Grok 4, and DeepSeek V3.2. Models were evaluated using identical inputs to eliminate prompt variability. We queried all models using their official UI interfaces to ensure stable and unmodified performance under standard usage conditions. Each model was accessed via its native interface - for instance, ChatGPT-5 through the OpenAI UI, Claude 4.5 Sonnet via the Anthropic UI, Gemini 2.5 Flash through Google AI Studio, and DeepSeek V3.2 via the DeepSeek UI interface. Grok 4 was accessed through the xAI UI.

The results generated by each model were automatically compared to the verified Omni Calculator output, providing a direct measure of computational correctness. Prompts were intentionally diverse in style - ranging from direct numerical queries to reverse or assumption-based formulations, and covered a balanced mix of conceptual reasoning and numeric computation. The dataset comprised a diverse range of problems, spanning from moderately difficult questions to more advanced, complex, and challenging ones, to evaluate the models under various conditions and scenarios. The most trivial prompts and situations were excluded, leaving sufficient room for error pattern analysis and longitudinal improvement tracking.

## 3.4. Scoring and Evaluation Framework

Model responses were assessed using a binary scoring system (1 = correct, 0 = incorrect), after normalizing units and rounding precision to match Omni's display format. Additional error categories were applied to qualify incorrect outputs, including:

1. **Formula or method error** – incorrect equation selection, e.g., using the wrong physical formula to find the kinetic energy of an object;
2. **Calculation error** – arithmetic or computational slip, e.g., $\sin 30° + \cos 60° = 0.866$;
3. **Unit or rounding error** – mismatched conversion or precision, e.g, 0.003839 N instead of 0.003827 N;
4. **Wrong assumption** – invalid or missing constant, e.g., using the universal gas constant without converting temperature to kelvins;
5. **Wrong parameters** – input values were incorrectly interpreted, omitted, or substituted during reasoning, e.g., treating a 5% annual rate as a monthly interest rate;
6. **Refusal/deflection** – safety refusal or generic advice instead of computing, e.g., declining to calculate due to a medical disclaimer;
7. **Hallucination** – invented data, constants, or references, e.g., fabricating an external reference instead of performing the actual calculation; or

8. **Incomplete or partial answer** – only part of the required output was produced, e.g., omitting the perimeter in an area-and-perimeter calculation.

In some cases, when a model's answer was incorrect, we tested its ability to self-correct after gentle feedback. For example, after an error in solving the equations ("5x+3y=1 and 6x+5y=5, solve it"), the model was prompted with "x is actually -1.629. Calculate it again, properly this time." These checks were excluded from scoring but offered additional insight into reasoning adaptability.

## 3.1. Units, Precision, and Scoring Normalization

All model outputs were standardized to Omni's canonical units for each calculator. Equivalent unit conversions (e.g., kWh ↔ kJ) were considered correct when numerically accurate. Scoring adhered to the display precision of Omni's results: an answer was marked correct only if the rounded value matched Omni's displayed output exactly. When models produced fewer decimals, they were prompted once to refine the output to the appropriate precision before scoring. The final example of used prompts, scoring, and evaluation method was presented in **Table 1**.

**Table 1.** Example of a benchmark prompt and model responses. Correct values are marked in green. Errors are classified according to the categories described in Section 3.5.

| Prompt | Omni's answer | ChatGPT 5 (free) | Gemini 2.5 Flash | Claude Sonnet 4.5 | Grok 4 | DeepSeek V3.2 |
|---|---|---|---|---|---|---|
| *If I deposit $50,000 at 5% APR, compounded weekly, what will my balance be after 18 months?* | $53892.27 | $53908.65 | $53892.27 | $53906.5 | $53892.27 | $53890.45 |
| **Scoring/Error category** | Reference | 0 (calculation error) | 1 | 0 (calculation error) | 1 | 0 (precision/rounding issue) |

# 4. Results

## 4.1. Overall accuracy of LLM Models

As previously noted, each model's output was evaluated using a binary scoring scheme (1 = response consistent with the calculator's verified result; 0 = incorrect). A total set of 500 prompts was distributed across 14 application domains corresponding to the functional scope of Omni calculators (**Table 2**).

**Table 2.** Distribution of 500 ORCA benchmark prompts across 13 of Omni's calculator-based categories, showing the relative representation of each thematic area used in model evaluation.

| Finance | 42 | Statistics | 34 |
|---|---|---|---|
| Math | 103 | Construction | 11 |
| Health | 60 | Sports | 10 |
| Other | 18 | Conversion | 5 |
| Biology | 42 | Everyday life | 12 |
| Physics | 146 | Ecology | 3 |
| Chemistry | 14 | | |

Specific prompt design and selection were carried out by subject-matter experts to ensure domain fidelity and conceptual rigor. Formally,

$$Accuracy = \frac{n_{correct}}{n_{total}}$$

where $n_{correct}$ is the number of correct answers for a given model, and $n_{total}$ is the total number of evaluated prompts with the model response obtained (each assigned a binary score of 0 or 1).

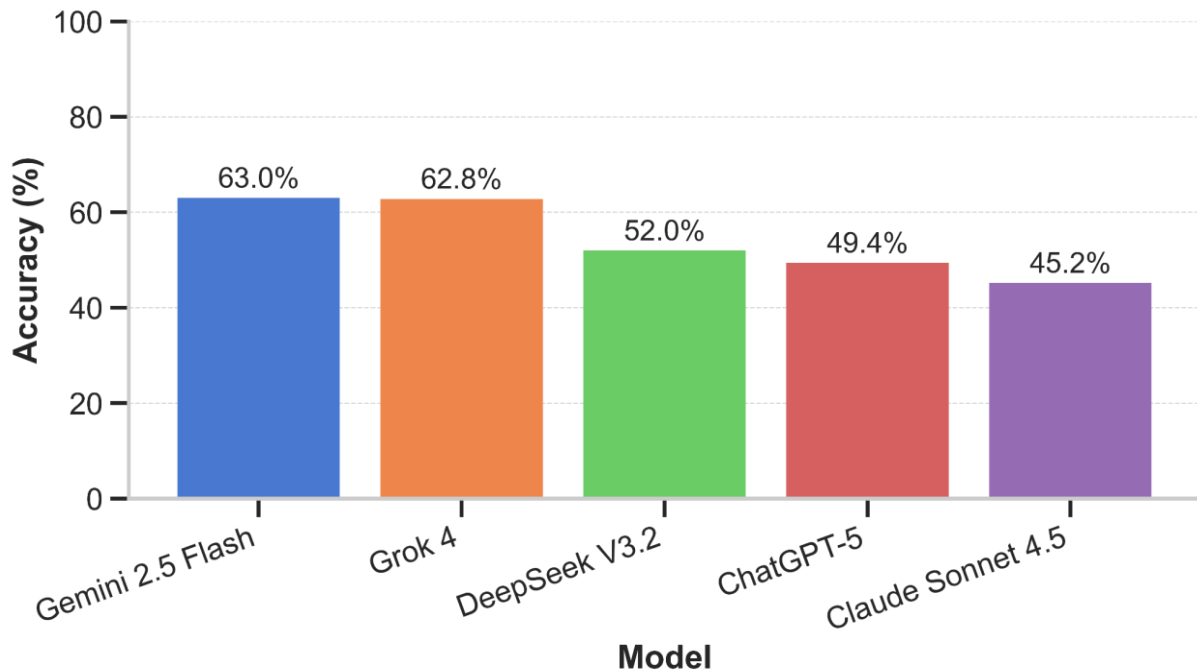

**Figure 1.** Overall accuracy of five state-of-the-art LLMs on the ORCA benchmark (n_total = 500 calculator-based tasks). Accuracy represents the percentage of correctly solved problems verified against Omni's deterministic computation engine.

**Figure 1** revealed that among the tested systems, Gemini 2.5 Flash achieved the highest overall accuracy (63%), followed closely by Grok 4 (62.8%), with DeepSeek V3.2 ranking third at 52.0%. ChatGPT-5 and Claude Sonnet 4.5 performed comparably but at lower levels (49.4% and 45.2%, respectively), indicating that even the most advanced proprietary models still fail on roughly half of all deterministic reasoning tasks. These results confirm that progress in natural-language reasoning does not directly translate into consistent computational reliability.

Despite clear gains compared to earlier LLM generations, the current models still fall short of the deterministic accuracy demonstrated by our computational benchmark, particularly in problems requiring intermediate assumptions or multistep reasoning.

## 4.2. Error Types Distribution Among Different LLMs

The distribution of error types provides deeper insight into the cognitive and computational weaknesses characteristic of current large language models. As we showed in **Figure 2**, the majority of incorrect responses originated from calculation errors (33.4%) and precision or rounding issues (34.7%), indicating that even when models follow an appropriate reasoning path, they frequently fail in the final numerical execution or rounding stage.

Together, these two categories account for over two-thirds of all observed errors, highlighting a persistent gap between linguistic reasoning and precise arithmetic manipulation.

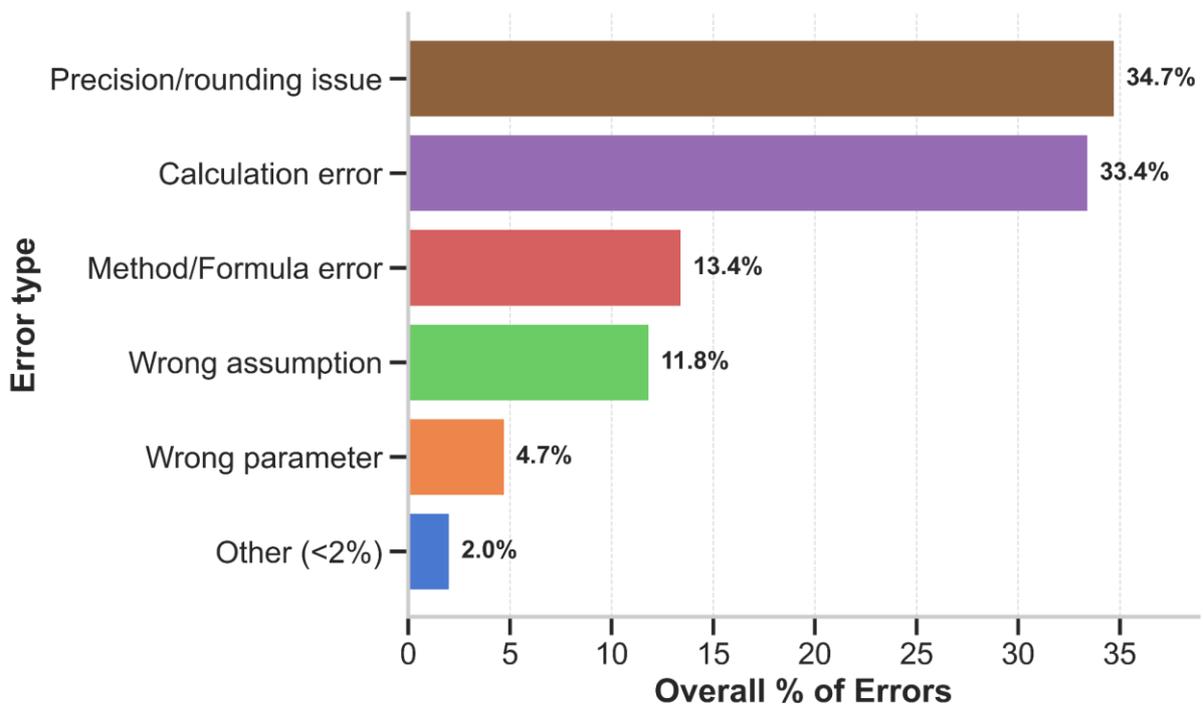

**Figure 2.** Overall distribution of error types across all evaluated LLMs. Colors represent distinct error categories, maintaining consistency with the categories introduced in Section 3.4. On-chart labels are displayed only for categories exceeding 2% of total errors.

A smaller but substantial portion of errors was classified as method or formula selection (13.4%) and wrong assumption (11.8%), both representing notable shares of the total error pool. Less frequent categories included incorrect parameter substitution (4.7%), while rare errors, such as incomplete or partial answers (1.3%), refusal or deflection (0.6%), hallucinations (0.1%), and unit errors (0.0%), were aggregated into the Other group in Figure 2 for visual clarity. Although these categories individually contribute marginally to the overall error rate, they collectively illustrate residual inconsistencies in model behavior and were therefore retained in the textual analysis for completeness.

Overall, the chart shows that five categories - calculation, precision or rounding, formula, assumption, and parameter errors - account for more than 97% of all incorrect outputs, indicating that the majority of LLM failures are concentrated at a limited number of recurring error types.

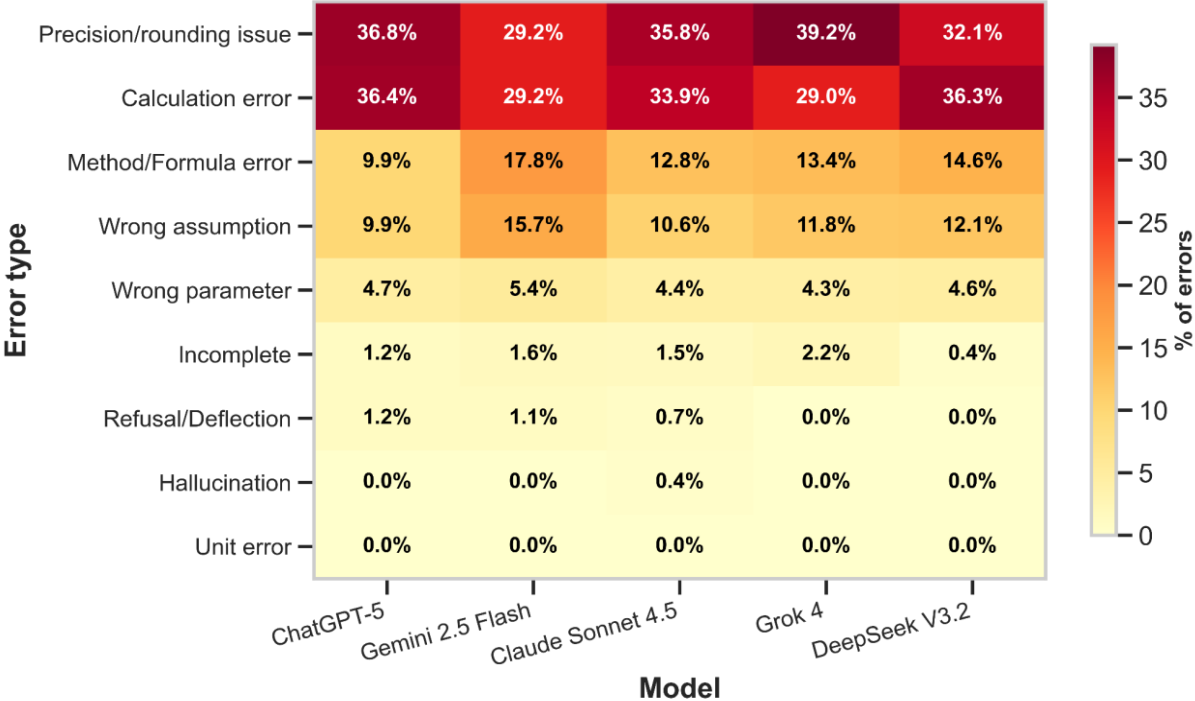

**Figure 3.** Heatmap showing the distribution of error types across the five evaluated LLMs. Color intensity represents the absolute number of errors per category, with darker shades indicating higher frequency. Labels inside each cell display raw error counts, while the yellow-to-red palette provides visual emphasis on magnitude differences.

The heatmap in Figure 3 was introduced to capture the proportional distribution of specific error types across all evaluated models. The majority of mistakes fall into two mechanical categories - precision or rounding issues (29.2–39.2%) and calculation errors (29–36.4%) - confirming that numerical instability remains the dominant limitation for all systems. While Gemini 2.5 Flash recorded the lowest relative share of both calculation and precision errors (≈29.2%), Grok 4 showed a mixed profile: with few calculation mistakes (29%) but the highest proportion of rounding issues (39.2%). In contrast, Claude Sonnet 4.5 and ChatGPT-5 showed consistently higher values in both categories, exceeding 33% each. Both models also produced a comparable number of method or formula errors and wrong assumptions, indicating similar overall error magnitude across the main reasoning-related categories. At the same time, Gemini 2.5 Flash showed the highest proportion of method or formula errors (17.8%) and wrong assumptions (15.7%), indicating a tendency toward conceptual rather than purely arithmetic mistakes.

In contrast, DeepSeek V3.2 displayed a mixed pattern, with relatively high counts in precision (32.1%) and calculation errors (36.3%) but noticeably fewer instances of interpretive mistakes such as wrong parameters (4.6%) or incomplete answers (0.4%). Across all models, refusals, hallucinations, and unit-related errors were negligible, each contributing ≤1% of total errors.

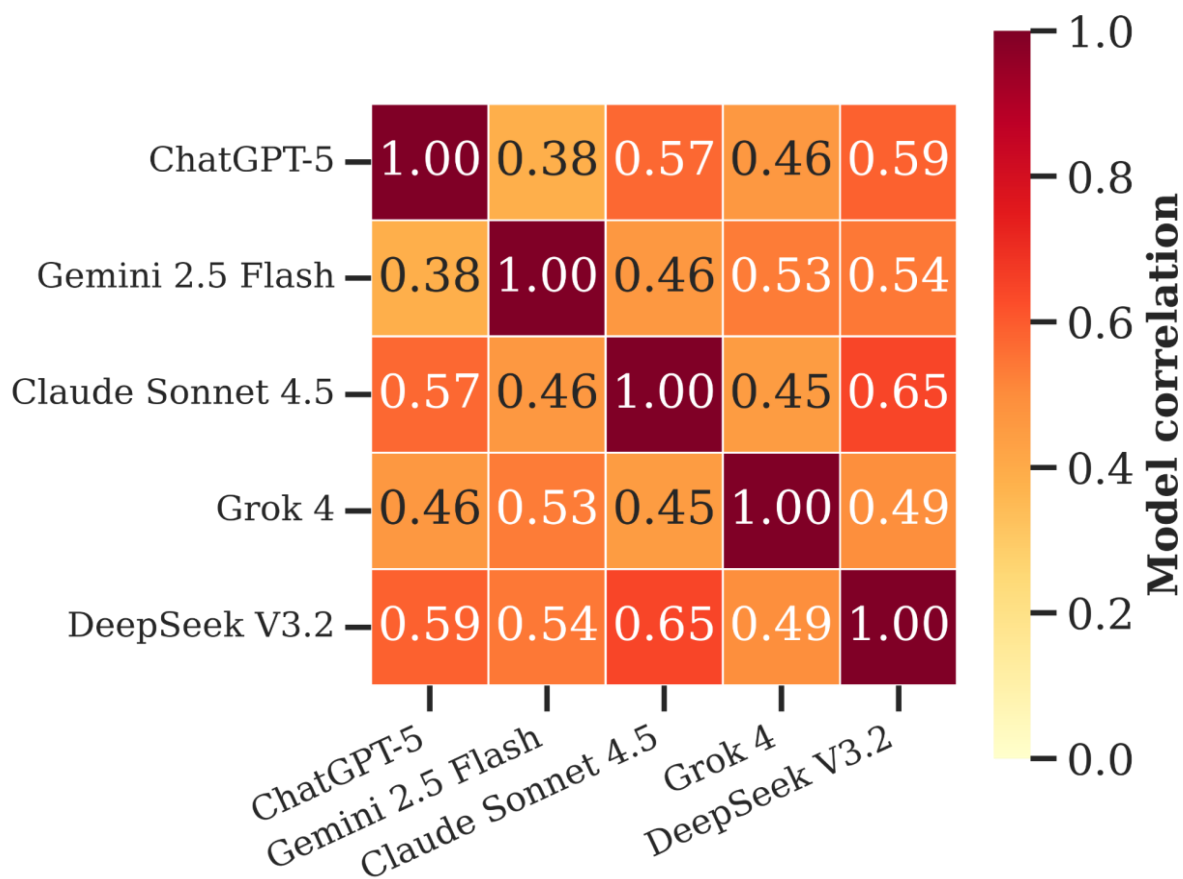

**Figure 4.** Correlation heatmap comparing the prompt-level answer patterns across all five evaluated LLMs.

In Figure 4, we collected responses from all 500 prompts across the evaluated models to compute a cross-model correlation matrix. Each cell represents the Pearson correlation coefficient between two models' binary correctness vectors (1 = correct, 0 = incorrect). Values close to 1 indicate that the models tend to succeed and fail on the same prompts, whereas lower values reflect more independent or divergent behavior.

Overall, the correlations range from 0.38 to 0.65, suggesting moderate overlap in model weaknesses and successes. The strongest association is observed between Claude Sonnet 4.5 and DeepSeek V3.2 (r = 0.65), implying that these two systems often produce similar outcomes on the same tasks. By contrast, Gemini 2.5 Flash exhibits lower correlations with most models (≈ 0.38–0.54), indicating a somewhat different reasoning profile and error distribution.

### 4.3. Best and Worst LLMs Across The Main Reasoning Domains

To facilitate clearer interpretation and enable meaningful cross-domain comparison, the original set of calculator categories was consolidated into seven principal domains reflecting major areas of applied quantitative reasoning: Biology & Chemistry, Mathematics & Conversions, Engineering & Construction, Finance & Economics, Health & Sports, Physics, and Statistics & Probability (**Table 3**). By merging conceptually overlapping or sparsely represented categories, we ensured that every domain contained a sufficient number of prompts to capture distinct reasoning patterns without fragmenting the data.

**Table 3.** Distribution of 500 ORCA benchmark prompts across the narrowed set of 7 main thematic domains, showing the relative representation of each thematic area used in model evaluation.

| Biology & Chemistry | 57 |
|---|---|
| Engineering & Construction | 27 |
| Finance & Economics | 43 |
| Health & Sports | 67 |
| Math & Conversions | 147 |
| Physics | 128 |
| Statistics & Probability | 31 |

In this way, we ensured that the resulting structure balances thematic coverage with analytical clarity-maintaining representativeness of real-world problem contexts while minimizing noise introduced by narrowly defined or low-frequency categories.

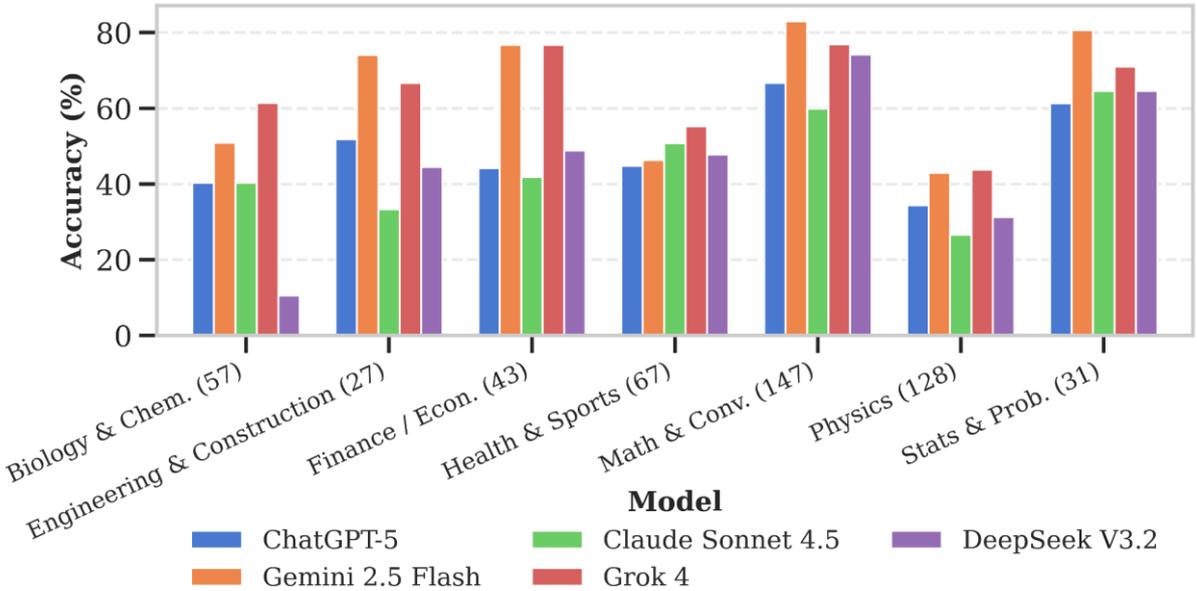

**Figure 5.** Accuracy of five LLMs across ten reasoning domains. Each bar represents the percentage of correctly solved prompts within a given domain. The domain names were slightly shortened on the x-axis (e.g., Biology & Chemistry → Biology & Chem.). Numbers in brackets indicate the number of prompts in each main domain.

Figure 5 shows that *Math & Conversions* and *Statistics & Probability* emerged as the highest-scoring domains, where several models, most notably Gemini 2.5 Flash (orange bar) and Grok 4 (red bar), achieved accuracy rates above 70%. In contrast, *Physics*, *Biology & Chemistry*, and *Health & Sports* yielded the lowest mean accuracies, with most models performing below 50%. *Finance & Economics*, and *Engineering & Construction* occupied an intermediate range, typically between 45% and 75% of correct answers for each model. Notably, *Gemini 2.5 Flash* demonstrated a distinct advantage in *Statistics & Probability*, achieving the highest accuracy (80.6%) among all models in this domain. The same model also reached an accuracy of 83% in the *Math & Conversions* category, further highlighting its strong quantitative reasoning capabilities.

When comparing models, Grok 4 and Gemini 2.5 Flash consistently ranked among the top performers across most domains, while Claude Sonnet 4.5 (green bar) and ChatGPT-5 (blue bar) demonstrated moderate but stable accuracy profiles. DeepSeek V3.2 (violet bar), on the other hand, exhibited the largest domain variability: performing strongly in *Math & Conversions*, yet dropping sharply in *Biology & Chemistry* (only 10.5% correct answers) and *Physics*.

**Table 4.** Accuracy of each model across seven main knowledge domains in ORCA benchmark (% of correct final answers).

| Main domain | ChatGPT-5 | Gemini 2.5 Flash | Claude Sonnet 4.5 | Grok 4 | DeepSeek V3.2 |
|---|---|---|---|---|---|
| **Biology & Chemistry** | 40.4% | 50.9% | 40.4% | 61.4% | 10.5% |
| **Engineering & Construction** | 51.9% | 74.1% | 33.3% | 66.7% | 44.4% |
| **Finance & Economics** | 44.2% | 76.7% | 41.9% | 76.7% | 48.8% |
| **Health & Sport** | 44.8% | 46.3% | 50.7% | 55.2% | 47.8% |
| **Math & Conversion** | 66.7% | 83.0% | 59.9% | 76.9% | 74.1% |
| **Physics** | 34.4% | 43.0% | 26.6% | 43.8% | 31.3% |
| **Statistics & Probability** | 61.3% | 80.6% | 64.5% | 71.0% | 64.5% |

Table 4 presents the detailed accuracy results for five leading AI models across seven main domains, as evaluated in the ORCA benchmark. Models with the highest accuracy in each domain were highlighted in green, while those with the lowest accuracy were marked in red. The table above highlights how model performance varies by subject.

The data also reveal differences in cross-model consistency within each domain. In areas such as *Math* and *Engineering & Construction*, the models' results cluster tightly, with no single system showing a clear outlier. By contrast, domains like *Biology / Life Science* and *Chemistry* display wider spreads - here, Grok 4 markedly outperforms DeepSeek V3.2 and ChatGPT-5, whose results fall to roughly one-third of Grok's accuracy in the same tasks. Similarly, *Physics* shows pronounced divergence: while Gemini 2.5 Flash and Grok 4 maintain moderate accuracy, Claude Sonnet 4.5 and DeepSeek V3.2 perform substantially worse.

# 5. Discussion

## 5.1. Overall Performance and Persistent Limitations

The overall accuracy of the evaluated models remains moderate, ranging from roughly 45% to 63%, with Gemini 2.5 Flash and Grok 4 consistently outperforming their counterparts. Although the differences are notable, they do not suggest a qualitative leap in computational reasoning. This pattern echoes prior findings from benchmarks such as GSM8K and MATH, which have repeatedly shown that even large-scale language models struggle with multi-step quantitative reasoning and symbolic arithmetic. Model scale and instruction tuning appear to improve linguistic fluency more than they enhance numerical reliability, leaving systematic calculation errors largely unresolved [6], [8].

### 5.2. Dominant Error Types

The most prevalent failure types were rounding or precision issues (34.7%) and calculation errors (33.4%), both of which reflect mechanical rather than conceptual weaknesses. This aligns with prior evidence that LLMs often produce arithmetically inconsistent results, even when their reasoning chain is sound. Several studies have shown that coupling a model with an external computational module - such as in Program-Aided Language Models [44] or Toolformer [45] - substantially mitigates these mechanical slips. The pattern observed here, therefore, underscores that the bottleneck lies not in the choice of reasoning path, but in the execution of numerically exact operations [29].

### 5.3. Domain-level Variability

Accuracy varied considerably across domains. The highest scores were observed in Mathematics & Conversions, and Probability & Statistics (typically 65–75%), while the lowest appeared in Physics and Health & Sports. This distribution suggests that tasks relying on straightforward arithmetic or unit transformations are handled more reliably than those requiring physical interpretation or cross-domain reasoning. In our experiments, DeepSeek V3.2 demonstrated strong specialization in computational and mathematical reasoning, performing well in numerically grounded problems but failing almost completely in Health & Sports. In contrast, Gemini 2.5 Flash and Grok 4 showed exceptional performance in the Finance & Economics domain, clearly outperforming the remaining models. Similar gradients were reported in the Minerva benchmark [46], where models exhibited strong symbolic competence but weaker applied understanding. This pattern supports the view that large language models inherit a strong directional bias from their pretraining objectives: systems like DeepSeek tend to optimize for deterministic and computation-heavy tasks, while models such as Gemini 2.5 Flash or Grok 4 exhibit enhanced performance in domains involving applied reasoning and real-world decision-making. In both cases, domain-focused pretraining appears to strengthen problem-solving efficiency within a specific class of reasoning while reducing flexibility and generalization across conceptual boundaries [7].

### 5.4. Cross-model Correlation of Failures

To further examine model diversity, we analyzed the prompt-level correlation of correctness across all 500 problems. The resulting 5×5 matrix revealed moderate to high pairwise correlations (r ≈ 0.38–0.65), indicating that the models often succeed and fail on overlapping subsets of prompts. However, no pair exhibited near-perfect alignment, suggesting that their reasoning failures are only partially shared. This pattern implies that each model maintains distinct inductive biases and error sensitivities, shaped by architectural and training differences. From a systems' perspective, such partial overlap is advantageous: ensembles or hybrid setups combining models with non-identical error profiles could yield greater robustness than any single system alone. The correlation analysis thus reinforces the view that architectural diversity contributes not only to performance variance but also to collective complementarity in reasoning coverage.

### 5.5. Error Consistency and Scarcity of Hallucinations

Interestingly, hallucinations and refusals were nearly absent (≤1%), which diverges from results obtained in open-ended question-answering tasks. The deterministic nature of the benchmark - each problem having a single, verifiable ground-truth value - likely constrained the generative freedom that typically leads to factual hallucination. Instead, errors clustered around measurable computational slips, reinforcing the view that numerical reasoning represents a distinct failure mode compared with linguistic generation [28].

### 5.6. Methodological Implications

Three implications emerge from these results. First, deterministic verification - through strict rounding and unit normalization - remains essential for the evaluation of reasoning accuracy. Second, in-context reasoning should not be conflated with mathematical correctness: even when a model produces a coherent "chain of thought" (CoT) approach, the final computation may diverge from the verified solution [47]. Combining CoT with symbolic or code-based execution appears to yield more stable results. Finally, the dominance of rounding and arithmetic errors supports recent claims that LLMs exhibit representation fragility: even small perturbations in numerical format or phrasing can alter the outcome [48].

### 5.7. Practical Takeaway

Considered together, these results reinforce a growing consensus: large language models can plan reasoning steps but cannot yet reliably compute - even in tasks grounded in everyday quantitative reasoning. Our ORCA benchmark, based on real-life calculator problems verified by Omni's computational engines, exposes this limitation directly: models can articulate logical procedures, yet they often falter in translating them into exact computation. Therefore, the most promising path forward lies in hybrid architectures - systems that use language models to decompose and explain problems while delegating numerical precision to dedicated computational backends.

## 6. Conclusions

We compared and evaluated five contemporary large language models (ChatGPT-5, Gemini 2.5 Flash, Claude 4.5 Sonnet, Grok 4, and DeepSeek V3.2) - using our newly developed ORCA (Omni Research on Calculation in AI) Benchmark grounded in real-life quantitative reasoning. By aligning 500 natural-language prompts with deterministic outputs from verified Omni calculator engines, we provide a uniquely transparent measure of each model's ability to transform verbal reasoning into computational accuracy.

The ORCA framework was specifically designed to address a persistent blind spot in language-model evaluation: the ability to perform real-world quantitative reasoning, rather than merely symbolic or competition-style mathematics. By grounding all tasks in authentic calculator-based problems and verifying each response against exact computational output, we aimed to assess how reliably current LLMs can reproduce the structured, step-by-step reasoning that underlies everyday problem-solving.

Even in every day, non-specialized problems, the overall accuracy of tested models did not exceed 63%, underscoring that the gap between verbal reasoning and verifiable computation persists despite significant architectural progress. This plateau suggests that while LLMs can simulate understanding through fluent explanations, they still lack the internal consistency and grounding necessary for dependable quantitative reasoning.

The results reveal a persistent gap between linguistic fluency and numerical reliability. Models such as Gemini 2.5 Flash and Grok 4 show high proficiency in applied and domain-specific reasoning, particularly in finance and everyday quantitative contexts, while DeepSeek V3.2 demonstrates strong performance in strictly computational and optimization-oriented tasks but fails to generalize across conceptual or interpretive domains. ChatGPT-5 and Claude Sonnet 4.5, though more balanced, exhibit similar magnitudes of mechanical and conceptual errors, underscoring the limits of current architectures in unifying symbolic reasoning with precise calculation.

At the same time, correlation analysis revealed only moderate overlap in error patterns ($r \approx 0.38–0.65$), indicating that while models often show comparable error rates, their failure profiles differ markedly: for instance, one tends to make methodological mistakes, whereas another shows more computational slips. The highest correlation was observed between Claude Sonnet 4.5 and DeepSeek V3.2, and the lowest ($\approx 0.38$) between ChatGPT-5 and Gemini 2.5 Flash. This partial independence suggests that architectural and training differences yield distinct error sensitivities, reinforcing the potential of hybrid or ensemble approaches to enhance robustness through complementary reasoning strategies.

Across all systems, the predominant failures stem from misapplied formulas, arithmetic slips, and misinterpreted parameters - errors that highlight the models' reliance on pattern-based heuristics rather than grounded computational logic. These findings reinforce a growing consensus: language models can plan reasoning steps but cannot yet perform them reliably, even at the level of everyday problem-solving.

## Disclaimer

This publication was reviewed by Hanna Pamuła, PhD, Head of Research and Content at Omni. Drawing on her expertise in machine learning and deep learning for signal processing, she ensured the technical accuracy and clarity of the content.

Final proofreading was carried out by Steven Wooding, Editorial Quality Assurance Coordinator at Omni, to maintain the highest standards of readability and consistency.

# 7. Appendix

## Sample prompts with model reply and error type classified

**1. Biology & Chemistry**

Prompt: What will the adult weight be in lbs of a 3 kg 6-week-old puppy?

Expected answer: 57.32 lb (51.59-63.05 lb)
DeepSeek V2.3: 26.32 lb

Error type: Wrong assumption

DeepSeek assumed that a puppy's weight is around 25% of the adult weight, instead of using the following equation: (puppy weight/puppy age in weeks) × 52.

**2. Engineering & Construction**

Prompt: Consider that you have 7 blue LEDs (3.6V) connected in parallel, together with a resistor, subject to a voltage of 12 V and a current of 5 mA. What is the value of the power dissipation in the resistor (in mW)?

Expected result: 42 mW
Claude Sonnet 4.5: 294 mW

Error type: Wrong parameter

→ Claude incorrectly applied 5 mA per LED instead of the total circuit current.

**3. Finance & Economics**

Prompt: I'm 35, plan to retire at 67, save $500/month, have $40,000 saved, and expect a 7% annual return (interest is compounded annually). How much will I have saved from now until retirement?

Expected result: $1,009,919.76
Grok 4:$1,056,211.38

→ Grok calculated 40,000 × (1.07)^32 ≈ $348,659 instead of $348,610.83.

**4. Health & Sports**

Prompt: What is my BMI if I am female, weigh 180 lbs and am 5'3?

Expected result: 31.9
Gemini 2.5 Flash: 31.8

Error type: Precision/rounding issue

→ Gemini gave a correct value (31.874) but then rounded down incorrectly.

### 5. Math & Conversions

Prompt: What is the area of a hexagram with side length 5 cm?

Expected result: 129.9 cm^2
DeepSeek V2.3: 21.65 cm^2

Error type: Method/Formula error

→ DeepSeek used the wrong formula for the hexagram's internal hexagon: $\sqrt{3}/2 \times a^2$ instead of $(3\sqrt{3}) \times a^2$.

### 6. Physics

Prompt: Consider that a metal composite has Poisson's ratio of 1.04, Young's modulus in x-direction equals to 78 MPa, Young's modulus in y-direction equals to 132 MPa, and shear modulus in xy-plane equals to 112 MPa. What will be the stress concentration factor of such metal composite? Write your answer with five significant figures.

Expected result: 1.3924
ChatGPT-5: It's not determinable from the information given.

Error type: Refusal/deflection

→ ChatGPT refused to answer, claiming insufficient data, even though the required formula can be derived from the given parameters.

### 7. Statistics & Probability

Prompt: For a lottery where 6 balls are drawn from a pool of 76, what are my chances of matching 5 of them?

Expected answer: 1 in 520521
ChatGPT-5: 1 in 401397

Error type: Calculation error

→ ChatGPT incorrectly computed 76C6 as 2,265,690,666 instead of 218,618,940.